\documentclass[letterpaper, 10 pt, conference, twoside]{ieeeconf} 
\pdfoutput=1
\usepackage{graphicx}
\usepackage{amsmath}
\usepackage{bbding}
\usepackage{pifont}
\usepackage{wasysym}
\usepackage{amssymb}
\usepackage{hyperref}
\usepackage{threeparttable} 
\usepackage{cite} 
\usepackage{booktabs}
\usepackage{xcolor}
\usepackage{multirow}
\usepackage{blindtext}
\usepackage{flushend}
\usepackage{float}  

\usepackage{fancyhdr}

\usepackage{amsmath,amssymb,amsfonts}

\usepackage{array} 

\usepackage{tikz,xcolor,hyperref}
\hypersetup{hidelinks}

\usepackage{stix} 
\definecolor{pothole}{RGB}{192, 64, 0}
\definecolor{potholeFP}{RGB}{253, 253, 100}
\definecolor{potholeFN}{RGB}{64, 128, 64}

\definecolor{road}{RGB}{64, 0, 128}
\definecolor{roadFP}{RGB}{0, 128, 192}
\definecolor{roadFN}{RGB}{128, 128, 0}

\usepackage{bm}
\usepackage{subcaption}
\usepackage[linesnumbered, ruled]{algorithm2e}

\SetKwInput{KwInput}{Input}
\SetKwInput
{KwOutput}{Output}

\usepackage{cite}

\IEEEoverridecommandlockouts  
\SetKwComment{Comment}{$\triangleright$\ }{}

\usepackage{afterpage}
\usepackage{mwe}

\title{\LARGE \bf Adaptive-Mask Fusion Network for Segmentation of Drivable Road and Negative Obstacle With Untrustworthy Features}
 
\author{Zhen Feng$^{1,2}$, Yuchao Feng$^{1}$, Yanning Guo$^{2}$, and Yuxiang Sun$^{1,*}$ 
\thanks{$^{1}$The Hong Kong Polytechnic University, Hung Hom, Kowloon, Hong Kong (email: zfeng94@outlook.com; yuchao.feng@connect.polyu.hk; yx.sun@polyu.edu.hk, sun.yuxiang@outlook.com).}
\thanks{$^{2}$Harbin Institute of Technology, Harbin, Heilongjiang, China (email: zfeng94@outlook.com; guoyn@hit.edu.cn).}
\thanks{$^*$Corresponding author: Yuxiang Sun.}}

\begin{document}

\maketitle
\thispagestyle{fancy}



\begin{abstract}
Segmentation of drivable roads and negative obstacles is critical to the safe driving of autonomous vehicles. Currently, many multi-modal fusion methods have been proposed to improve segmentation accuracy, such as fusing RGB and depth images. However, we find that when fusing two modals of data with untrustworthy features, the performance of multi-modal networks could be degraded, even lower than those using a single modality. In this paper, the untrustworthy features refer to those extracted from regions (e.g., far objects that are beyond the depth measurement range) with invalid depth data (i.e., $0$ pixel value) in depth images. The untrustworthy features can confuse the segmentation results, and hence lead to inferior results. To provide a solution to this issue, we propose the Adaptive-Mask Fusion Network (AMFNet) by introducing adaptive-weight masks in the fusion module to fuse features from RGB and depth images with inconsistency. In addition, we release a large-scale RGB-depth dataset with manually-labeled ground truth based on the NPO dataset for drivable roads and negative obstacles segmentation. Extensive experimental results demonstrate that our network achieves state-of-the-art performance compared with other networks. Our code and dataset are available at: \href{https://github.com/lab-sun/AMFNet}{https://github.com/lab-sun/AMFNet}.
\end{abstract}

\section{Introduction}

Segmentation of drivable roads and negative obstacles is a fundamental capability for autonomous vehicles. Although vehicles can generally pass small negative obstacles on roads, negative obstacles are still potential threats to vehicles. Especially when vehicle speed is fast or negative obstacles are large, severe accidents, such as roll over, could happen \cite{thegood}. 
Accurate segmentation results of drivable roads and negative obstacles could serve as input data for downstream tasks, such as path planning \cite{cheng2022sequential}, to avoid potential accidents.

Many single-modal (e.g., using only RGB images) networks have been proposed for the segmentation of drivable roads and negative obstacles \cite{orsic2019defense,pereira2019semantic}. To improve the segmentation performance, multi-modal networks based on RGB-depth (RGB-D) fusion \cite{feng2022mafnet,wang2021dynamic,fan2020roadseg} and RGB-disparity fusion \cite{fan2020we, wang2021dynamic} have been proposed. 
Although these networks have achieved acceptable results, we find that when there are a large number of pixels in depth images without valid depth information (i.e., pixel value $0$ in depth images), the segmentation performance cannot been improved or even inferior to the performance with a single RGB modality. We call the regions with the pixel value $0$ in depth images as untrusted regions.
The value $0$ in depth images indicates that the depth information of the object cannot be measured (e.g., out of the depth measurement range), rather than indicating that the distance between the object and the camera is $0$. The features extracted from untrusted regions could not represent the real features of the environment, so we call the features as untrustworthy features. The untrustworthy features could confuse the segmentation results since there are valid features from the other modality. 

To provide a solution to this issue, we propose a novel Adaptive-Mask Fusion Network (AMFNet) with Adaptive-Mask Fusion (AMF) modules. To this end, we generate mask images from depth images to distinguish trusted and untrusted regions. The AMF module is used to generate adaptive-weight masks for RGB and depth feature maps to reduce the influence caused by untrustworthy features during fusion. We also release a large-scale RGB-depth (RGB-D) dataset with manually-labeled ground truth for drivable roads and negative obstacles segmentation. Our contributions are summarized as follows: 
\begin{itemize}
    \item We propose an Adaptive-Mask Fusion (AMF) module to reduce the influence of untrustworthy features during feature fusion.
    \item We proposed a novel fusion network named AMFNet with the AMF modules for the segmentation of drivable roads and negative obstacles.
    \item We release a large-scale RGB-D dataset based on the NPO dataset\footnote{\href{https://github.com/lab-sun/InconSeg}{https://github.com/lab-sun/InconSeg}}. Our dataset consists of $8,752$ RGB-D images with manually-labeled ground truth for the segmentation of drivable roads and negative obstacles.
\end{itemize}

\section{Related Works} 
\subsection{Semantic Segmentation Networks}

Chen \textit{et al.} \cite{chen2018encoder} designed DeepLabV3+ with atrous convolution in encoder-decoder structure for semantic segmentation. 
Azad \textit{et al.} \cite{azad2020attention} introduced attention modules into DeepLabV3+ to propose Att-Deeplabv3+. Recently, many Transformer-based semantic segmentation networks have been proposed. 
Hatamizadeh \textit{et al.} \cite{hatamizadeh2022swin} combined the U-shaped structure and Transformer structure to design Swin UNETR for medical image segmentation. 
Yuan \textit{et al.} \cite{yuan2023effective} proposed CTC-Net for medical image segmentation with a convolutional neural networks-based encoder and a transformer-based encoder. 

To improve semantic segmentation accuracy, many multi-modal fusion networks have been proposed. 
Hazirbas \textit{et al.} \cite{hazirbas2016fusenet} proposed FuseNet to fuse RGB images and depth images for semantic segmentation.
Sun \textit{et al.} \cite{sun2019rtfnet} proposed RTFNet to fuse RGB images and thermal images for the segmentation of urban scenes.
Zhou \textit{et al.} \cite{zhou2022frnet} proposed FRNet to fuse RGB images and depth images with a cross-level enriching module in the encoder.

\subsection{Semantic Segmentation of Drivable Road}

Wang \textit{et al.} \cite{wang2020applying} proposed a normal inference module (NIM) for the depth image to improve the performance of drivable areas and road anomaly detection. The performance of several networks embedded with NIM has been improved. 
Fan \textit{et al.} \cite{fan2020roadseg} proposed a surface normal estimator for depth images and designed RoadSeg to fuse the output of the surface normal estimator and RGB images. 
Kothandaraman \textit{et al.} \cite{kothandaraman2021sssfda} proposed an unsupervised method to segment roads under adverse weather conditions. 
Chin \textit{et al.} \cite{min2022orfd} proposed transformer-based OFF-Net to fuse LiDAR point cloud and RGB image. They also released the ORFD dataset with $12,198$ pairs of LiDAR point cloud and RGB images.

\subsection{Semantic Segmentation of Negative Obstacles}

Fan \textit{et al.} \cite{fan2020we} proposed AA-RTFNet by combining RTFNet and attention modules to fuse RGB images and disparity images. They also released the Pothole-600 dataset for the segmentation of potholes.
Feng \textit{et al.} \cite{feng2022mafnet} proposed MAFNet to fuse RGB images and disparity images for the segmentation of potholes.
Masihullah \textit{et al.} \cite{Masihullah2021Attention} combined attention modules with DeepLabV3+ to segment roads and potholes.

Although the aforementioned multi-modal fusion networks have achieved acceptable results, they all ignore the influence of untrustworthy features. We find that untrustworthy features could degrade the fusion performance. So, in this work, we propose the AMF module to reduce the influence of untrustworthy features.

\section{The Proposed Method} 

\subsection{The Overall Architecture}
Fig.~\ref{overall} shows the overall architecture of our proposed AMFNet. Our AMFNet is designed with the 
structure. It consists of a five-stage RGB encoder, a five-stage depth encoder, and a five-stage decoder. There are also 5 AMF modules in the RGB encoder. Each AMF module is placed behind each stage of the RGB encoder. The RGB encoder and depth encoder are borrowed from BotNet-50 \cite{srinivas2021botnet}. Depth images are used to generate masks with a threshold of $0$. When the value of a pixel in the depth image is greater than $0$, the value of the pixel in the mask is $1$. We believe that the value $0$ in the depth image is untrustworthy because the distance between the real environment point and the camera is not $0$. The mask is a map used to distinguish trustworthy pixels from untrustworthy pixels. The mask is downsampled to generate five different masks (i.e., $M_1$, $M_2$, $M_3$, $M_4$, and $M_5$), which have the same resolution as the outputs of the $5$ stages of the RGB encoder. The $M_n$ mask is fed into the $n$-th AMF module, where $n \in [0,5]$. The outputs of the same level stages of the RGB and depth encoders are fed into the same level AMF module. The output of the $n$-th AMF module is fed into the $(n+1)$-th stage of the RGB encoder and fused into the output of the $(n+1)$-th stage of the decoder by element-wise addition. 

\begin{figure*}[t] 
\setlength{\abovecaptionskip}{0pt} 
\setlength{\belowcaptionskip}{0pt} 
\renewcommand\arraystretch{1.0} 
\renewcommand\tabcolsep{7pt} 
	\centering 
	\includegraphics[width=7.0in]{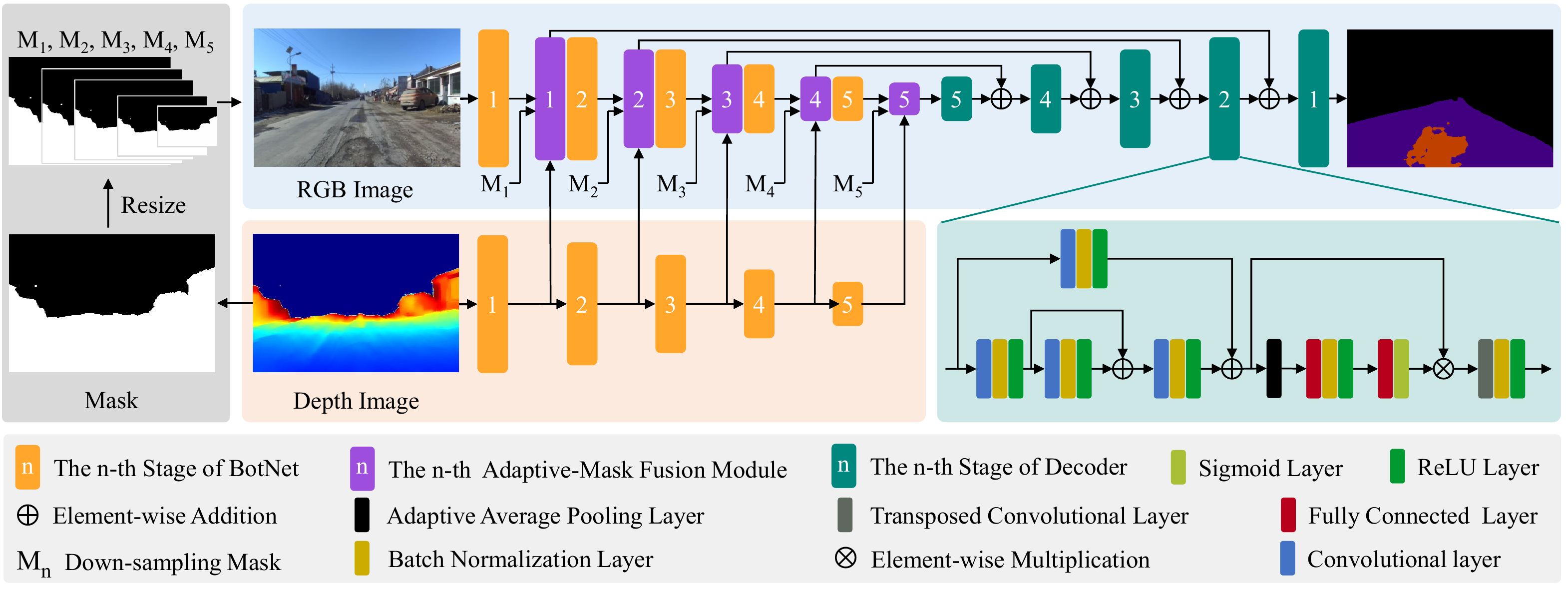} 
	\caption{The overall architecture of our proposed AMFNet. Our AMFNet adopts the two-encoder-one-decoder structure: a five-stage RGB encoder, a five-stage depth encoder, and a five-stage decoder. The encoder is adopted from BotNet-50 \cite{srinivas2021botnet}. Our proposed adaptive-mask fusion (AMF) modules are placed behind each stage of the RGB encoder. The mask is generated by thresholding the depth image with the pixel value $0$. Five different masks (i.e., $M_1$, $M_2$, $M_3$, $M_4$, and $M_5$) with the same resolution as the outputs of the $5$ stages of the RGB encoder are generated by downsampling with the nearest neighbor method. The figure is best viewed in color.}
	\label{myOverall}
\end{figure*}

\begin{figure}[t] 
\setlength{\abovecaptionskip}{0pt} 
\setlength{\belowcaptionskip}{0pt} 
\renewcommand\arraystretch{1.0} 
\renewcommand\tabcolsep{7pt} 
	\centering 
	\includegraphics[width=3.4in]{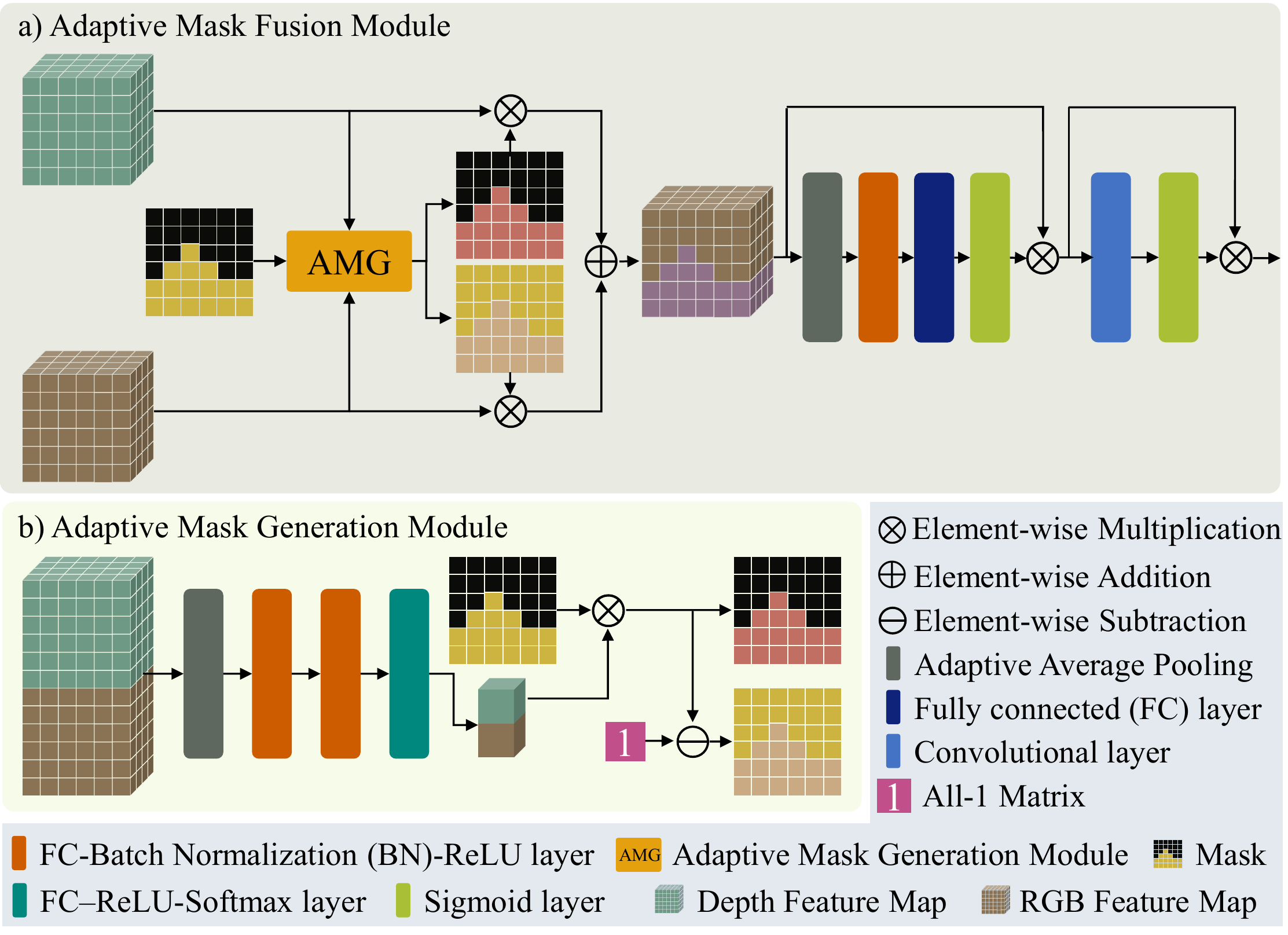} 
	\caption{The structure of an AMF module. Both AMF and adaptive mask generation (AMG) Module have the same three inputs: an RGB feature map, a depth feature map, and a mask. The outputs of the AMG are two adaptive-weight masks for the RGB feature map and depth feature map.}
	\label{AMFM}
\end{figure}
 
\subsection{The AMF Module} 
The structure of the AMF module is shown in Fig.~\ref{AMFM}. The $n$-th AMF module has three inputs: the output of the $n$-th stage of the RGB encoder (RGB feature map), the output of the $n$-th stage of the depth encoder (depth feature map), and the $n$-th mask $M_n$. In each AMF module, the mask is first fed into an adaptive mask generation (AMG) module to generate two adaptive-weight masks for the RGB feature map and depth feature map. Secondly, the adaptive-weight masks are fused with the RGB feature map and depth feature map by element-wise multiplication, and then fused to generate the result of the fusion of the RGB feature map and depth feature map by element-wise addition. Finally, the weights of the fusion result of the RGB feature map and depth feature map are adjusted by a channel attention block and a spatial attention block. In the channel attention block, the fusion result is passed through an adaptive average pooling layer, a fully connected (FC)-batch normalization (BN)-ReLU layer, an FC layer, and a Sigmoid layer sequentially to generate the weights of each channel. The weights of each channel are fused into the result of the fusion to generate an adjusted result by element-wise multiplication. In the spatial attention block, the adjusted result is passed through a convolutional layer and a Sigmoid layer sequentially to generate spatial weights. The spatial weights are fused into the adjusted result to generate the output of AMF module by element-wise multiplication.

The main purpose of the AMF module is to divide the feature map into trusted regions and untrusted regions according to the mask. In the trusted regions, RGB features and trustworthy depth features are fused by adaptive weights. In the untrusted regions, the untrustworthy features of the depth images are discarded, and the RGB features are directly used as the fusion result. The AMG module is designed to achieve this purpose. The AMG module has three inputs: the mask, the RGB feature map, and the depth feature map. The three inputs have the same resolution. Firstly, the RGB feature map and depth feature map are concatenated together. Secondly, the concatenated feature map is passed through an adaptive average pooling layer, two FC-BN-ReLU layers, and an FC-BN-Softmax layer. The outputs of the FC-BN-Softmax layer are two weights for the RGB feature map and depth feature map. Thirdly, the weight for the depth feature map is fused with the mask by element-wise multiplication to generate the mask for the depth feature map. Finally, the depth-feature-map mask is subtracted from an all-one matrix to generate the RGB-feature-map mask.

\subsection{The Decoder}

The decoder consists of five stages with the same structure. Fig.~\ref{overall} shows the structure of one stage. The input of one stage is first fed into a dual residual block. Secondly, the output of the dual residual block is fed into a channel attention block to adjust the weights of each channel. Finally, a transposed Convolution-BN-ReLU (CBR) layer is used to generate the output of the stage. 

There are four CBR layers in the dual residual block. The input is fed into the first CBR layer and the fourth CBR layer. The output of the first CBR layer is fed into the second CBR layer and fused with the output of the second CBR layer. The fusion result is fed into the third CBR layer. The outputs of the third and the fourth CBR layer are fused together as the output of the dual residual block.

\section{Experimental Results and Discussions}
\subsection{The Dataset}

As aforementioned, large-scale multi-modal datasets with drivable roads and negative obstacles are very limited. Therefore, we release a large-scale RGB-D dataset based on the NPO dataset for the segmentation of drivable roads and negative obstacles. The raw images of the NPO dataset were recorded with a ZED stereo camera mounted on a vehicle. There are $20$ image sequences in the raw data of the NPO dataset. We manually label one image per $5$ images in some image sequences that include nearly $44,000$ image pairs (left images, right images, and depth images) with $1,242 \times 2,208$ resolution. So, in total, $8,752$ images are labelled in our dataset. To alleviate the annotation task, we directly use the masks of negative obstacles in the NPO dataset as the masks for our annotation. We name our dataset as Drivable Roads and Negative Obstacles (DRNO) dataset. There are various lighting conditions, weather conditions, and scenes in our dataset, such as normal lighting, large areas of shadow, snowy, sunny, cloudy, urban scenes, and rural scenes. There are also various road surface types in the data set, such as water, snow, and normal road surfaces.

\begin{figure}[t] 
\renewcommand\arraystretch{1.0} 
\centering  
\subfloat{ 
 \centering  
 \begin{tikzpicture}
 \node[anchor=south west,inner sep=0] (image) at (0,0) {\includegraphics[width=0.1555\textwidth]{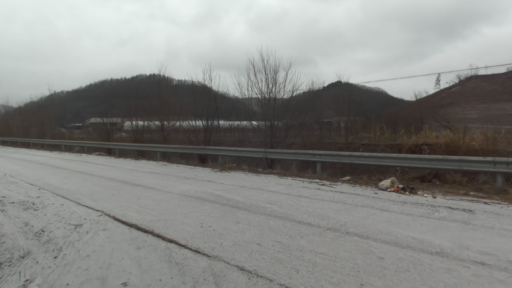}};
 \end{tikzpicture}
 \label{f11} 
 } \hspace{-0.32cm}
\subfloat{ 
 \centering 
 \begin{tikzpicture}
 \node[anchor=south west,inner sep=0] (image) at (0,0) {\includegraphics[width=0.1555\textwidth]{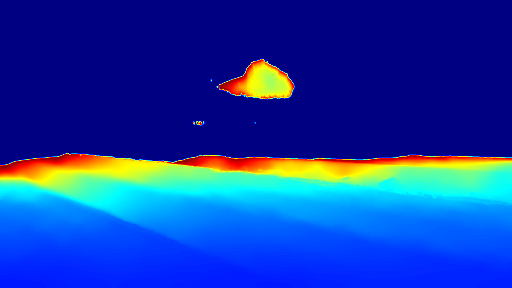}};
 \end{tikzpicture}
 \label{f12} 
 } \hspace{-0.32cm}
\subfloat{  
 \centering 
 \begin{tikzpicture}
 \node[anchor=south west,inner sep=0] (image) at (0,0) {\includegraphics[width=0.1555\textwidth]{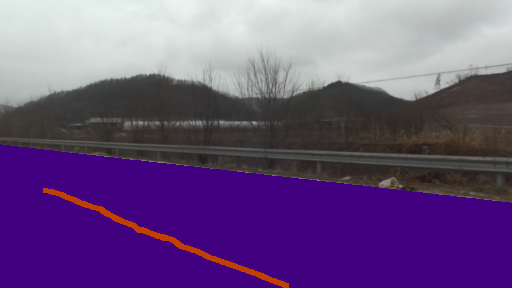}};
 \end{tikzpicture}
 \label{f13} 
 } 
\vspace{0.05cm}
\subfloat{ 
 \centering 
 \begin{tikzpicture}
 \node[anchor=south west,inner sep=0] (image) at (0,0) {\includegraphics[width=0.1555\textwidth]{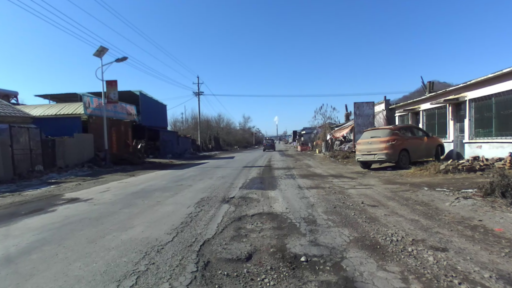}};
 \end{tikzpicture}
 \label{f21} 
 }  \hspace{-0.32cm}
\subfloat{ %
 \centering 
 \begin{tikzpicture}
 \node[anchor=south west,inner sep=0] (image) at (0,0) {\includegraphics[width=0.1555\textwidth]{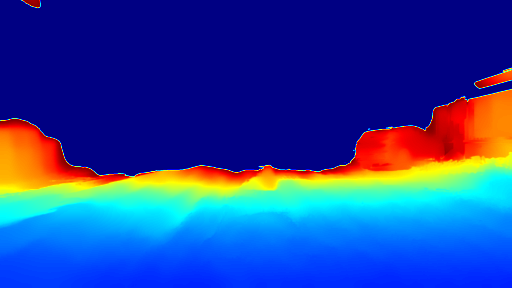}};
 \end{tikzpicture}
 \label{f22} 
 } \hspace{-0.32cm}
\subfloat{  
 \centering  
 \begin{tikzpicture}
 \node[anchor=south west,inner sep=0] (image) at (0,0) {\includegraphics[width=0.1555\textwidth]{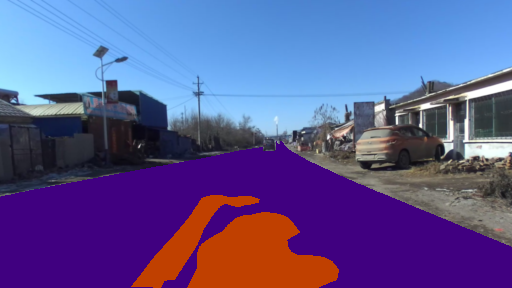}};
 \end{tikzpicture}
 \label{f23} 
 }
\vspace{0.05cm}
\subfloat{  
 \centering  
 \begin{tikzpicture}
 \node[anchor=south west,inner sep=0] (image) at (0,0) {\includegraphics[width=0.1555\textwidth]{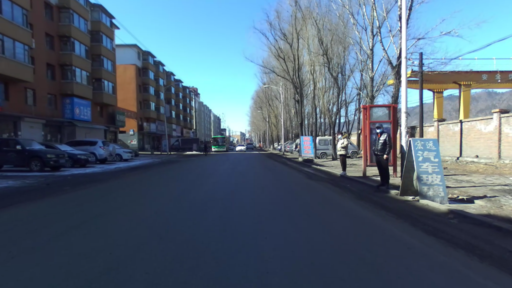}};
 \end{tikzpicture}
 \label{f31} 
 } \hspace{-0.32cm}
\subfloat{  
 \centering  
 \begin{tikzpicture}
 \node[anchor=south west,inner sep=0] (image) at (0,0) {\includegraphics[width=0.1555\textwidth]{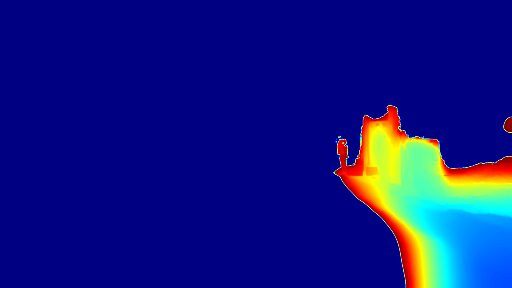}};
 \end{tikzpicture}
 \label{f32} 
 } \hspace{-0.32cm}
\subfloat{ 
 \centering 
 \begin{tikzpicture}
 \node[anchor=south west,inner sep=0] (image) at (0,0) {\includegraphics[width=0.1555\textwidth]{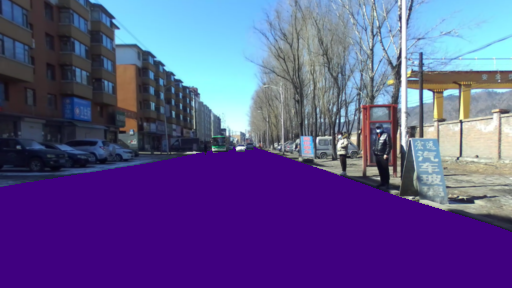}};
 \end{tikzpicture}
 \label{f33} 
 } 
\caption{Sample images in our DRNO dataset. We visualize the depth images with the \textit{jet} color map. Depth values increase from red to blue. \textcolor{road}{$\mdblksquare$} and \textcolor{pothole}{$\mdblksquare$} represent drivable roads and negative obstacles, respectively. The figure is best viewed in color.}  
\label{Sample}  
\end{figure}

To the best of our knowledge, our DRNO dataset is the largest dataset for semantic segmentation of drivable roads and negative obstacles. Some samples of our dataset are shown in Fig.~\ref{Sample}. In our DRNO dataset, $8,752$ images include drivable roads, and $748$ images include negative obstacles. 

\subsection{Training Details}

Our AMFNet is implemented with PyTorch. The network is trained and tested on a PC with NVIDIA RTX 3090 (24 GB RAM) graphics card. The parameters of the first four encoder stages of AMFNet are initialized with the pre-trained weight from PyTorch. We employ the stochastic gradient descent (SGD) optimizer to train the network. The initial learning rate is set to $0.01$, the momentum is set to $0.9$, and the decay strategy is set to $0.95$.

We randomly split our dataset into three sets: training ($4,376$ pairs of RGB-D images), validation ($2,188$ pairs of RGB-D images), and testing ($2,188$ pairs of RGB-D images). To trade-off training speed and network performance, we reduce the resolution of the images to $288 \times 512$ during training and testing. The unlabelled background is treated as a class during training and testing. 

\subsection{Ablation Study}

We conduct ablation study to check the benefits of the AMF module and choose the optimal structure for our AMFNet. Firstly, we place one AMF module behind the different stages of the RGB encoder to design variants. For example, an AMF module is placed behind the $5$-th stage of the RGB encoder to fuse RGB and depth feature maps in a variant. Secondly, we design variants by placing AMF modules behind different stages of the RGB encoder. For example, AMF modules are placed behind the last two stages of the RGB encoder in a variant. In all the variants, the outputs of the same-level stages without AMF modules are fused by element-wise addition. The metrics Mean Accuracy (mAcc), mean F-score (mF1), and mean Intersection-over-Union (mIoU) \cite{feng2022mafnet} are used to evaluate the performance of all the variants. 

The results are displayed in Tab.~\ref{abalationforstructure}. From the results, we can find that the variant without any AMF module achieves inferior results. Comparing variant A to variant F, we can find that no matter where AMF is placed, it can always improve the performance of the network. Comparing the four variants G, H, I, and J, more AMF modules in one variant lead to better performance. This shows that our proposed AMF module can remove untrustworthy features in the fusion process, thus improving the fusion performance. Based on the experimental results, we place five AMF modules behind each stage of the RGB encoder in our AMFNet.

\begin{table}[t]
 \setlength{\abovecaptionskip}{0pt} 
\setlength{\belowcaptionskip}{0pt} 
\renewcommand\arraystretch{1.0} 
\renewcommand\tabcolsep{7.6pt} 
\centering 
\begin{threeparttable}
\caption{The results (\%) of the ablation study. '\checkmark' means AMF module is placed behind the $n$-th stage of the RGB encoder. '$-$' means that the outputs of the $n$-th stage of the RGB encoder and depth encoder are fused with element-wise addition. The best results are highlighted in bold font.}
\begin{tabular}{@{}ccccccccc@{}}
    \toprule
    \multirow{1}[4]{*}{No.} & \multicolumn{5}{c}{Stage}             & \multicolumn{1}{c}{\multirow{1}[4]{*}{mAcc}} & \multicolumn{1}{c}{\multirow{1}[4]{*}{mIoU}} & \multicolumn{1}{c}{\multirow{1}[4]{*}{mF1}} \\
\cmidrule{2-6}          & 1st   & 2nd   & 3rd   & 4th   & 5th   &       &       &  \\
    \midrule
    (A)   & $-$ & $-$ & $-$  & $-$  & $-$  & 67.57  & 66.54  & 69.18  \\
    (B)   & $-$ & $-$  &  $-$   & $-$   & \checkmark    & 69.96 & 67.00  & 69.18 \\
    (C)   & $-$ &   $-$  &   $-$    & \checkmark &  $-$  & 68.36 & 67.05  & 69.87 \\
    (D)   & $-$ &  $-$  & \checkmark&   $-$ & $-$ & 68.48  & 66.86  & 69.72  \\
    (E)   & $-$ & \checkmark  &  $-$    &    $-$   &    $-$   &   67.79    & 66.58      & 69.22 \\
    (F)   & \checkmark & $-$  &  $-$ & $-$  & $-$   & 67.99  & 66.76  & 69.49  \\
\midrule
    (G)   &  $-$  & $-$  & $-$  & \checkmark  & \checkmark  & 68.89  & 67.21  & 70.15  \\
    (H)   &  $-$     &  $-$     & \checkmark & \checkmark & \checkmark & 69.10  &  67.27 &  70.24\\
    (I)   &   $-$    & \checkmark & \checkmark & \checkmark & \checkmark &  69.98     &  67.99     & 71.34 \\
    (J)   & \checkmark & \checkmark & \checkmark &\checkmark & \checkmark & \textbf{70.60} & \textbf{68.39}  & \textbf{71.99}  \\
    \bottomrule
    \end{tabular}
\label{abalationforstructure}
\end{threeparttable}
\end{table}

\begin{table*}[t] 
\setlength{\abovecaptionskip}{0pt} 
\setlength{\belowcaptionskip}{0pt} 
\renewcommand\arraystretch{1.0} 
\renewcommand\tabcolsep{2.5pt} 
\centering 
\begin{threeparttable}
\caption{The comparative results (\%) on the testing set of our DRNO dataset. 'Modality' means the type of modality for training networks. 'Year' means the published year of networks. 'RGB \& Depth' means the network is trained and tested with RGB-depth fusion modality. The results demonstrate the superiority of our AMFNet. The best results are highlighted in bold font.} 
\begin{tabular*}{\hsize}{@{}@{\extracolsep{\fill}}lcccccccccccccc@{}}
   \toprule
    \multirow{2}[4]{*}{Network} & \multirow{2}[4]{*}{Year} & \multirow{2}[4]{*}{Modality} & \multicolumn{3}{c}{Background} & \multicolumn{3}{c}{Drivable Road}   & \multicolumn{3}{c}{Negative Obstacles} & \multirow{2}[4]{*}{mAcc} & \multirow{2}[4]{*}{mIoU} & \multirow{2}[4]{*}{mF1} \\
\cmidrule(lr){4-6}  \cmidrule(lr){7-9} \cmidrule(lr){10-12}         &       &       & Acc   & IoU   & F1    & Acc   & IoU   & F1    & Acc   & IoU   & F1    &       &       &  \\
    \midrule
    \multirow{3}[2]{*}{FuseNet \cite{hazirbas2016fusenet}} & \multirow{3}[2]{*}{2016} & Depth & 98.50  & 97.40  & 98.68  & 97.67  & 94.61  & 97.23  & 0.00  & 0.00  & 0.00  & 65.39  & 64.00  & 65.30  \\
          &       & RGB   & 98.76  & 97.67  & 98.82  & 97.70  & 95.15  & 97.51  & 0.00  & 0.00  & 0.00  & 65.49  & 64.27  & 65.45  \\
          &       & RGB \& Depth & 98.95  & 97.59  & 98.78  & 97.13  & 94.96  & 97.42  & 0.00  & 0.00  & 0.00  & 65.36  & 64.18  & 65.40  \\
    \midrule
    \multirow{3}[2]{*}{RTFNet \cite{sun2019rtfnet}} & \multirow{3}[2]{*}{2019} & Depth & 99.20  & 97.65  & 98.81  & 96.71  & 95.02  & 97.45  & 1.39  & 1.28  & 2.52  & 65.77  & 64.65  & 66.26  \\
          &       & RGB   & 99.48  & 97.92  & 98.95  & 96.69  & 95.56  & 97.73  & 7.82  & 6.25  & 11.77  & 68.00  & 66.58  & 69.48  \\
          &       & RGB \& Depth & 99.50  & 97.90  & 98.94  & 96.56  & 95.47  & 97.68  & 7.07  & 4.21  & 8.08  & 67.71  & 65.86  & 68.23  \\
    \midrule
    \multirow{3}[2]{*}{AA-RTFNet \cite{fan2020we}} & \multirow{3}[2]{*}{2020} & Depth & 99.11  & 97.62  & 98.80  & 96.84  & 94.98  & 97.43  & 1.25  & 1.07  & 2.11  & 65.73  & 64.56  & 66.11  \\
          &       & RGB   & 99.47  & 98.13  & 99.05  & 97.06  & 95.93  & 97.92  & 14.14  & 7.11  & 13.27  & 70.22  & 67.06  & 70.08  \\
          &       & RGB \& Depth & 99.45  & 98.06  & 99.02  & 97.01  & 95.81  & 97.86  & 7.11  & 4.37  & 8.38  & 67.86  & 66.08  & 68.42  \\
    \midrule
    \multirow{3}[2]{*}{RoadSeg \cite{fan2020roadseg}} & \multirow{3}[2]{*}{2020} & Depth & 98.30  & 96.44  & 98.19  & 96.02  & 92.66  & 96.19  & 0.00  & 0.00  & 0.00  & 64.78  & 63.03  & 64.79  \\
          &       & RGB   & 98.92  & 98.19  & 99.09  & \textbf{98.44} & 96.18  & 98.06  & 8.39  & 6.83  & 12.78  & 68.58  & 67.07  & 69.97  \\
          &       & RGB \& Depth & 99.28  & 98.09  & 99.04  & 97.44  & 95.91  & 97.91  & 7.34  & 4.84  & 9.24  & 68.02  & 66.28  & 68.73  \\
    \midrule
    \multirow{3}[2]{*}{SS-SFDA \cite{kothandaraman2021sssfda}} & \multirow{3}[2]{*}{2021} & Depth & 98.47  & 96.62  & 98.28  & 96.05  & 93.00  & 96.37  & 0.00  & 0.00  & 0.00  & 64.84  & 63.21  & 64.88  \\
          &       & RGB   & 98.59  & 97.07  & 98.51  & 96.76  & 93.91  & 96.86  & 0.84  & 0.79  & 1.57  & 65.40  & 63.92  & 65.65  \\
          &       & RGB \& Depth & 98.61  & 96.89  & 98.42  & 96.34  & 93.54  & 96.66  & 0.07  & 0.07  & 0.14  & 65.01  & 63.50  & 65.08  \\
    \midrule
    \multirow{3}[2]{*}{MAFNet \cite{feng2022mafnet}} & \multirow{3}[2]{*}{2022} & Depth & 98.57  & 96.95  & 98.45  & 96.55  & 93.67  & 96.73  & 0.00  & 0.00  & 0.00  & 65.04  & 63.54  & 65.06  \\
          &       & RGB   & 99.44  & 98.14  & 99.06  & 97.26  & 96.04  & 97.98  & 8.11  & 7.35  & 13.69  & 68.27  & 67.17  & 70.24  \\
          &       & RGB \& Depth & 99.51  & 98.14  & 99.06  & 97.09  & 96.01  & 97.97  & 4.90  & 4.03  & 7.76  & 67.17  & 66.06  & 68.26  \\
    \midrule
    \multirow{3}[2]{*}{OFF-Net \cite{min2022orfd}} & \multirow{3}[2]{*}{2022} & Depth & 96.71  & 91.95  & 95.81  & 89.34  & 83.57  & 91.05  & 0.00  & 0.00  & 0.00  & 62.02  & 58.51  & 62.29  \\
          &       & RGB   & 98.17  & 96.60  & 98.27  & 96.66  & 93.02  & 96.38  & 0.00  & 0.00  & 0.00  & 64.94  & 63.21  & 64.88  \\
          &       & RGB \& Depth & 97.51  & 96.37  & 98.15  & 97.56  & 92.68  & 96.20  & 0.00  & 0.00  & 0.00  & 65.02  & 63.02  & 64.78  \\
    \midrule
    \multirow{3}[2]{*}{FRNet \cite{zhou2022frnet}} & \multirow{3}[2]{*}{2022} & Depth & 99.10  & 97.76  & 98.87  & 97.18  & 95.29  & 97.59  & 0.00  & 0.00  & 0.00  & 65.43  & 64.35  & 65.49  \\
          &       & RGB   & 99.42  & 97.92  & 98.95  & 96.83  & 95.58  & 97.74  & 6.64  & 5.89  & 11.13  & 67.63  & 66.47  & 69.27  \\
          &       & RGB \& Depth & \textbf{99.52} & 98.02  & 99.00  & 96.78  & 95.72  & 97.81  & 7.46  & 4.62  & 8.83  & 67.92  & 66.12  & 68.55  \\
    \midrule
    \multicolumn{2}{c}{\multirow{3}[2]{*}{AMFNet (Ours)}} & Depth & 99.07  & 97.58  & 98.78  & 96.85  & 94.90  & 97.39  & 1.08  & 0.99  & 1.97  & 65.67  & 64.49  & 66.04  \\
    \multicolumn{2}{c}{} & RGB   & 99.26  & 98.25  & 99.12  & 97.88  & 96.30  & 98.12  & 8.86  & 8.01  & 14.82  & 68.67  & 67.52  & 70.69  \\
    \multicolumn{2}{c}{} & RGB \& Depth & 99.25  & \textbf{98.40} & \textbf{99.19} & 98.17  & \textbf{96.57} & \textbf{98.26} & \textbf{14.39} & \textbf{10.20} & \textbf{18.51} & \textbf{70.60} & \textbf{68.39} & \textbf{71.99} \\
    \bottomrule
\end{tabular*} 
\label{overall}
\end{threeparttable}
\end{table*}
  
\subsection{Comparative Study}
We compare our proposed AMFNet with the well-known networks: FuseNet \cite{hazirbas2016fusenet}, RTFNet \cite{sun2019rtfnet}, AA-RTFNet \cite{fan2020we}, RoadSeg \cite{fan2020roadseg}, SS-SFDA \cite{kothandaraman2021sssfda}, MAFNet \cite{feng2022mafnet}, OFF-Net \cite{min2022orfd}, and FRNet \cite{zhou2022frnet}.
We use our DRNO dataset to train and test the networks. To illustrate the impact of untrustworthy features on existing multi-modal fusion networks, we also train and test the aforementioned multi-modal networks without RGB encoders or depth encoders. We also removed the fusion module from the multi-modal network during training and testing. In other words, each multi-modal fusion network is trained and tested by three different modalities, namely single RGB modality, single depth modality, and RGB-depth fusion modality. The mAcc, mF1, and mIoU are also used as metrics to evaluate the performance of our AMFNet and these networks. In addition, the Acc, F1, and IoU of each class (i.e., background, drivable road, and negative obstacles) are also used as evaluation metrics. 

\subsubsection{Quantitative Results}
The results of all networks are displayed in Tab.~\ref{overall}. Comparing the results of each multi-modal fusion network, we can find that the results of the single RGB modality of all networks are better than the results of the RGB-depth fusion modality, except our network. These networks fuse untrustworthy features as general features, thus degrading the results of multi-modal fusion. Our proposed network reduces the influences of the untrustworthy features through the AMF module, making the results of the RGB-depth fusion modality better than that of the single-RGB modality. These results confirm our conjecture that untrustworthy features hinder multi-modal fusion. Comparing all the results, our network almost achieves the best results in terms of all metrics. This illustrates the superiority of our AMFNet.
   
\subsubsection{Qualitative Demonstrations}
\begin{figure*}[t] 
\renewcommand\arraystretch{1.0} 
\centering 
\subfloat{  
 \centering 
 \begin{tikzpicture}
 \node[anchor=south west,inner sep=0] (image) at (0,0) {\includegraphics[width=0.122\textwidth]{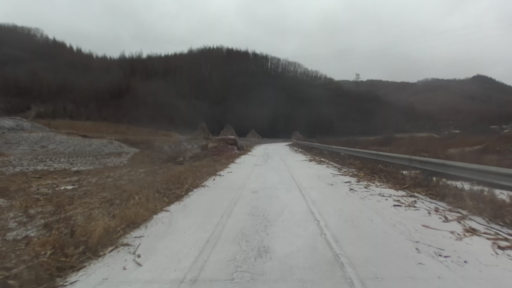}};
 \end{tikzpicture}
 \label{f011} 
 } \hspace{-0.33cm}
\subfloat{  
 \centering 
 \begin{tikzpicture}
 \node[anchor=south west,inner sep=0] (image) at (0,0) {\includegraphics[width=0.122\textwidth]{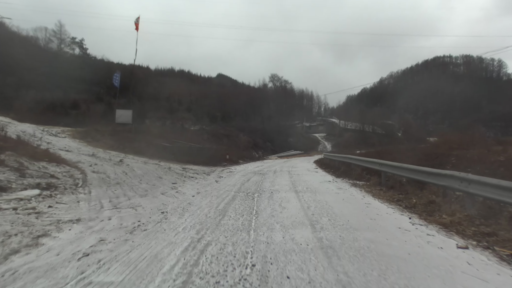}};
 \end{tikzpicture}
 \label{f012} 
 } \hspace{-0.33cm}
\subfloat{  
 \centering 
 \begin{tikzpicture}
 \node[anchor=south west,inner sep=0] (image) at (0,0) {\includegraphics[width=0.122\textwidth]{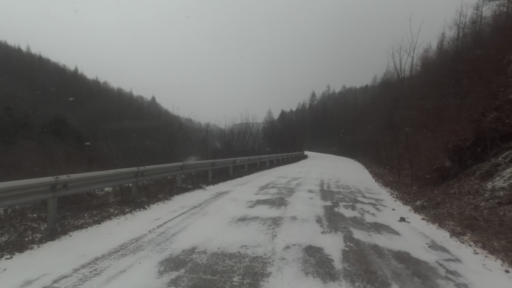}};
 \end{tikzpicture}
 \label{f013} 
 } \hspace{-0.33cm}
\subfloat{  
 \centering 
 \begin{tikzpicture}
 \node[anchor=south west,inner sep=0] (image) at (0,0) {\includegraphics[width=0.122\textwidth]{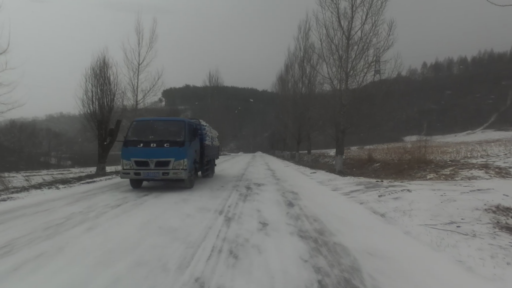}};
 \end{tikzpicture}
 \label{f014} 
 } \hspace{-0.33cm}
\subfloat{  
 \centering 
 \begin{tikzpicture}
 \node[anchor=south west,inner sep=0] (image) at (0,0) {\includegraphics[width=0.122\textwidth]{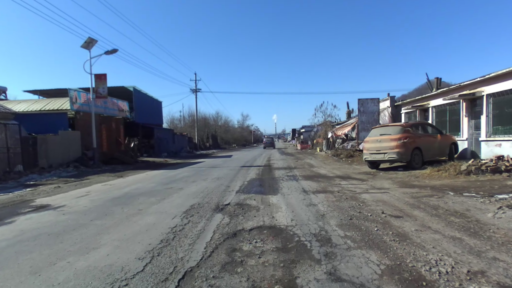}};
 \end{tikzpicture}
 \label{f015} 
 } \hspace{-0.33cm}
\subfloat{  
 \centering 
 \begin{tikzpicture}
 \node[anchor=south west,inner sep=0] (image) at (0,0) {\includegraphics[width=0.122\textwidth]{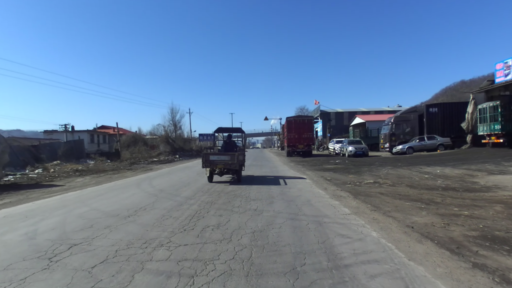}};
 \end{tikzpicture}
 \label{f016} 
 } \hspace{-0.33cm}
\subfloat{  
 \centering 
 \begin{tikzpicture}
 \node[anchor=south west,inner sep=0] (image) at (0,0) {\includegraphics[width=0.122\textwidth]{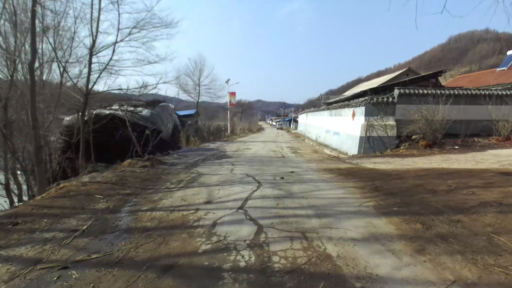}};
 \end{tikzpicture}
 \label{f017} 
 } \hspace{-0.33cm}
\subfloat{  
 \centering 
 \begin{tikzpicture}
 \node[anchor=south west,inner sep=0] (image) at (0,0) {\includegraphics[width=0.122\textwidth]{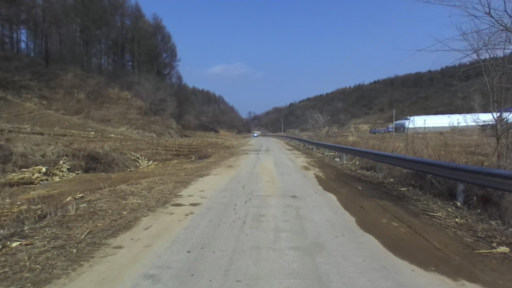}};
 \end{tikzpicture}
 \label{f018} 
 }

\vspace{0.02cm}

\subfloat{ %
 \centering 
 \begin{tikzpicture}
 \node[anchor=south west,inner sep=0] (image) at (0,0) {\includegraphics[width=0.122\textwidth]{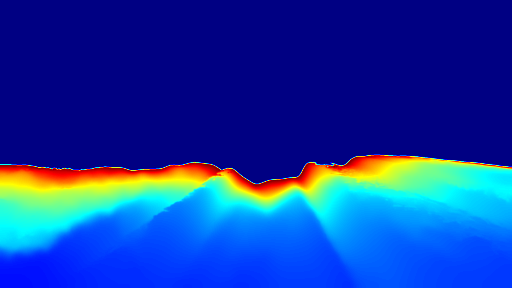}};
 \end{tikzpicture}
 \label{f021} 
 } \hspace{-0.33cm}
\subfloat{  
 \centering 
 \begin{tikzpicture}
 \node[anchor=south west,inner sep=0] (image) at (0,0) {\includegraphics[width=0.122\textwidth]{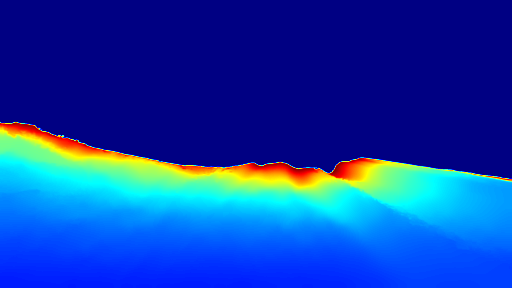}};
 \end{tikzpicture}
 \label{f022} 
 } \hspace{-0.33cm}
\subfloat{  
 \centering 
 \begin{tikzpicture}
 \node[anchor=south west,inner sep=0] (image) at (0,0) {\includegraphics[width=0.122\textwidth]{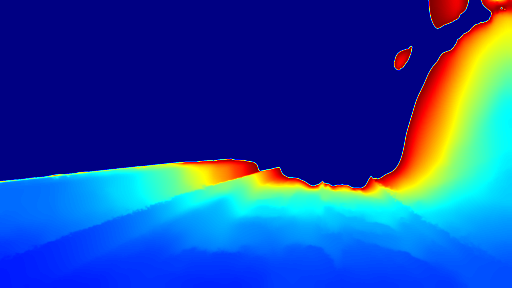}};
 \end{tikzpicture}
 \label{f023} 
 } \hspace{-0.33cm}
\subfloat{  
 \centering 
 \begin{tikzpicture}
 \node[anchor=south west,inner sep=0] (image) at (0,0) {\includegraphics[width=0.122\textwidth]{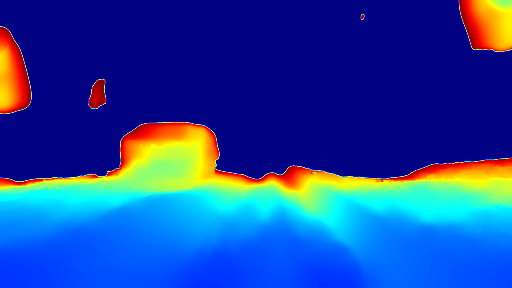}};
 \end{tikzpicture}
 \label{f024} 
 } \hspace{-0.33cm}
\subfloat{  
 \centering 
 \begin{tikzpicture}
 \node[anchor=south west,inner sep=0] (image) at (0,0) {\includegraphics[width=0.122\textwidth]{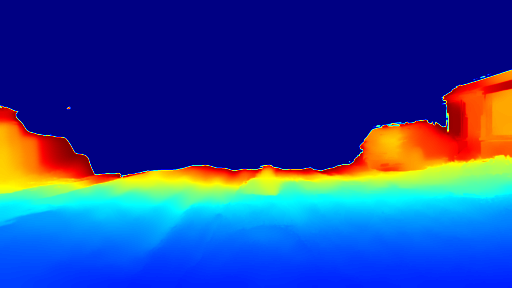}};
 \end{tikzpicture}
 \label{f025} 
 } \hspace{-0.33cm}
\subfloat{  
 \centering 
 \begin{tikzpicture}
 \node[anchor=south west,inner sep=0] (image) at (0,0) {\includegraphics[width=0.122\textwidth]{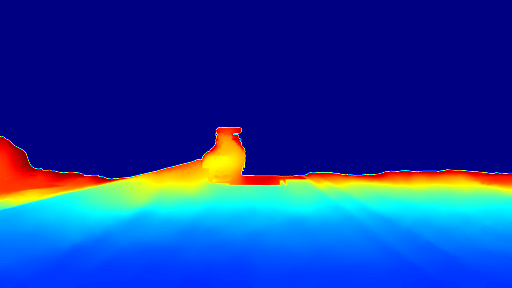}};
 \end{tikzpicture}
 \label{f026} 
 } \hspace{-0.33cm}
\subfloat{  
 \centering 
 \begin{tikzpicture}
 \node[anchor=south west,inner sep=0] (image) at (0,0) {\includegraphics[width=0.122\textwidth]{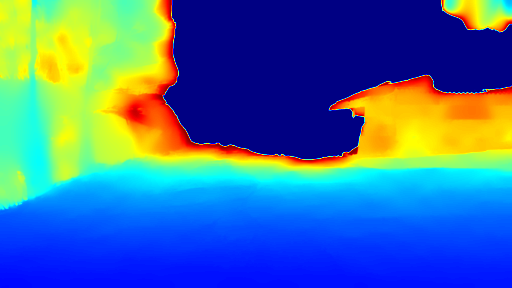}};
 \end{tikzpicture}
 \label{f027} 
 } \hspace{-0.33cm}
\subfloat{  
 \centering 
 \begin{tikzpicture}
 \node[anchor=south west,inner sep=0] (image) at (0,0) {\includegraphics[width=0.122\textwidth]{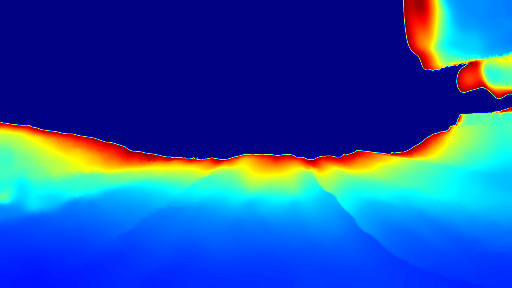}};
 \end{tikzpicture}
 \label{f028} 
 }

\vspace{0.02cm}

\subfloat{ %
 \centering 
 \begin{tikzpicture}
 \node[anchor=south west,inner sep=0] (image) at (0,0) {\includegraphics[width=0.122\textwidth]{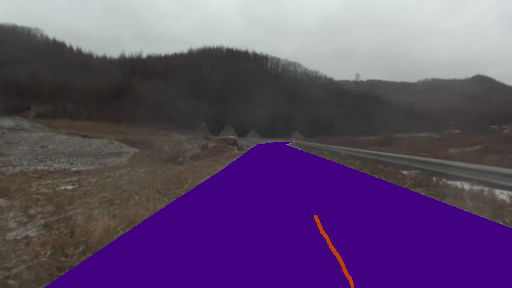}};
 \end{tikzpicture}
 \label{f031} 
 } \hspace{-0.33cm}
\subfloat{  
 \centering 
 \begin{tikzpicture}
 \node[anchor=south west,inner sep=0] (image) at (0,0) {\includegraphics[width=0.122\textwidth]{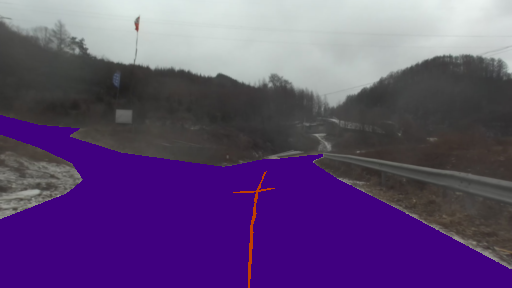}};
 \end{tikzpicture}
 \label{f032} 
 } \hspace{-0.33cm}
\subfloat{  
 \centering 
 \begin{tikzpicture}
 \node[anchor=south west,inner sep=0] (image) at (0,0) {\includegraphics[width=0.122\textwidth]{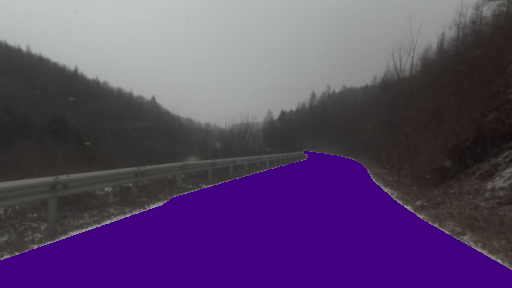}};
 \end{tikzpicture}
 \label{f033} 
 } \hspace{-0.33cm}
\subfloat{  
 \centering 
 \begin{tikzpicture}
 \node[anchor=south west,inner sep=0] (image) at (0,0) {\includegraphics[width=0.122\textwidth]{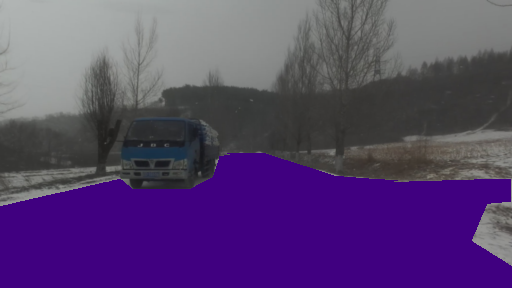}};
 \end{tikzpicture}
 \label{f034} 
 } \hspace{-0.33cm}
\subfloat{  
 \centering 
 \begin{tikzpicture}
 \node[anchor=south west,inner sep=0] (image) at (0,0) {\includegraphics[width=0.122\textwidth]{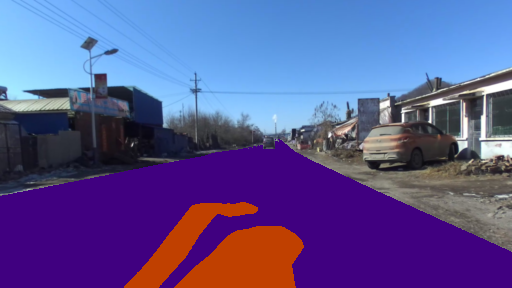}};
 \end{tikzpicture}
 \label{f035} 
 } \hspace{-0.33cm}
\subfloat{  
 \centering 
 \begin{tikzpicture}
 \node[anchor=south west,inner sep=0] (image) at (0,0) {\includegraphics[width=0.122\textwidth]{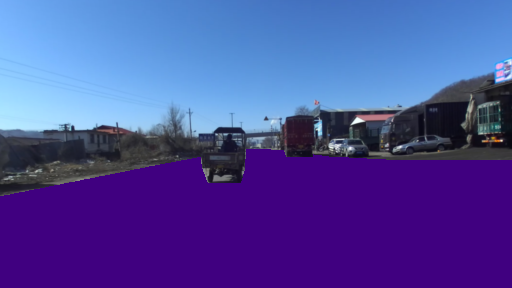}};
 \end{tikzpicture}
 \label{f036} 
 } \hspace{-0.33cm}
\subfloat{  
 \centering 
 \begin{tikzpicture}
 \node[anchor=south west,inner sep=0] (image) at (0,0) {\includegraphics[width=0.122\textwidth]{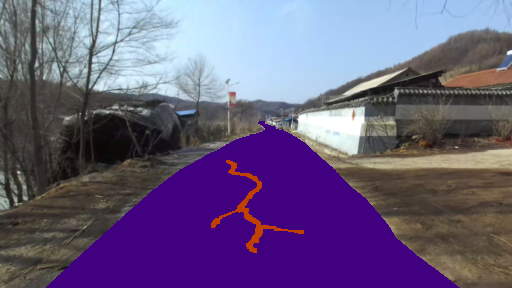}};
 \end{tikzpicture}
 \label{f037} 
 } \hspace{-0.33cm}
\subfloat{  
 \centering 
 \begin{tikzpicture}
 \node[anchor=south west,inner sep=0] (image) at (0,0) {\includegraphics[width=0.122\textwidth]{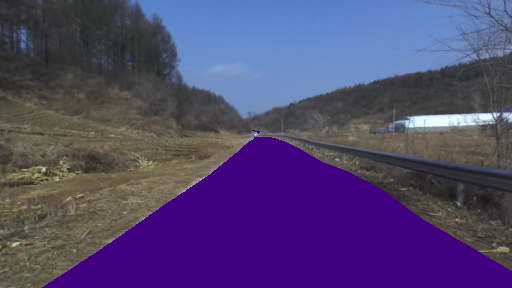}};
 \end{tikzpicture}
 \label{f038} 
 }

 \vspace{0.02cm}

\subfloat{ 
 \centering 
 \begin{tikzpicture}
 \node[anchor=south west,inner sep=0] (image) at (0,0) {\includegraphics[width=0.122\textwidth]{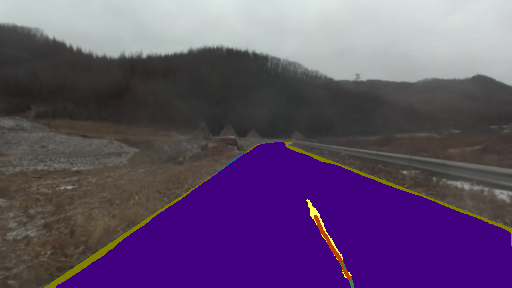}};
 \end{tikzpicture}
 \label{f041} 
 } \hspace{-0.33cm}
\subfloat{  
 \centering 
 \begin{tikzpicture}
 \node[anchor=south west,inner sep=0] (image) at (0,0) {\includegraphics[width=0.122\textwidth]{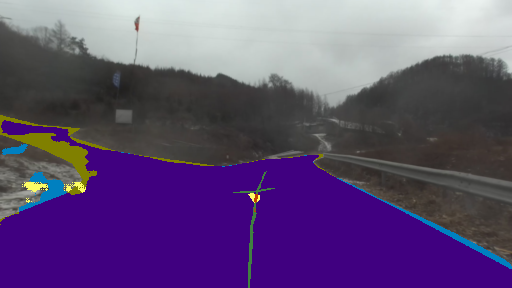}};
 \end{tikzpicture}
 \label{f042} 
 } \hspace{-0.33cm}
\subfloat{  
 \centering 
 \begin{tikzpicture}
 \node[anchor=south west,inner sep=0] (image) at (0,0) {\includegraphics[width=0.122\textwidth]{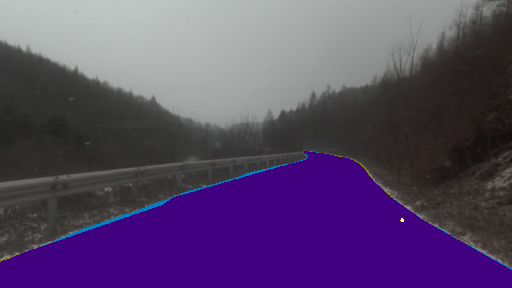}};
 \end{tikzpicture}
 \label{f043} 
 } \hspace{-0.33cm}
\subfloat{  
 \centering 
 \begin{tikzpicture}
 \node[anchor=south west,inner sep=0] (image) at (0,0) {\includegraphics[width=0.122\textwidth]{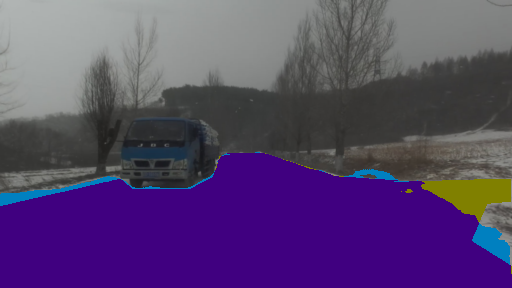}};
 \end{tikzpicture}
 \label{f044} 
 } \hspace{-0.33cm}
\subfloat{  
 \centering 
 \begin{tikzpicture}
 \node[anchor=south west,inner sep=0] (image) at (0,0) {\includegraphics[width=0.122\textwidth]{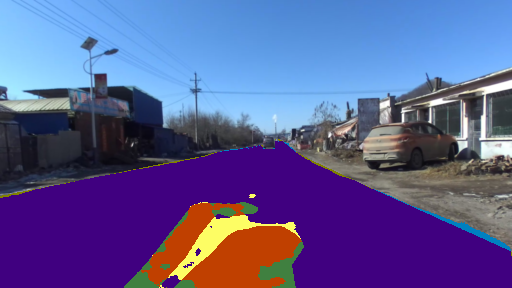}};
 \end{tikzpicture}
 \label{f045} 
 } \hspace{-0.33cm}
\subfloat{  
 \centering 
 \begin{tikzpicture}
 \node[anchor=south west,inner sep=0] (image) at (0,0) {\includegraphics[width=0.122\textwidth]{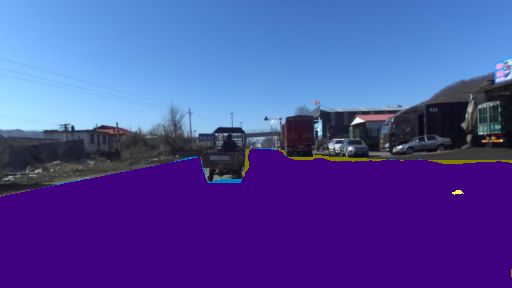}};
 \end{tikzpicture}
 \label{f046} 
 } \hspace{-0.33cm}
\subfloat{  
 \centering 
 \begin{tikzpicture}
 \node[anchor=south west,inner sep=0] (image) at (0,0) {\includegraphics[width=0.122\textwidth]{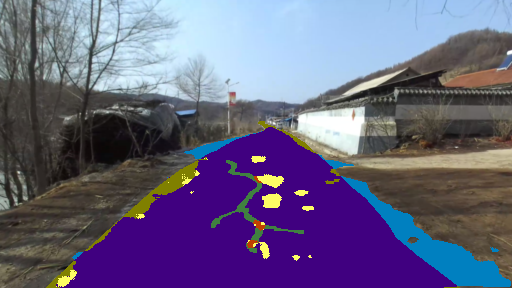}};
 \end{tikzpicture}
 \label{f047} 
 } \hspace{-0.33cm}
\subfloat{  
 \centering 
 \begin{tikzpicture}
 \node[anchor=south west,inner sep=0] (image) at (0,0) {\includegraphics[width=0.122\textwidth]{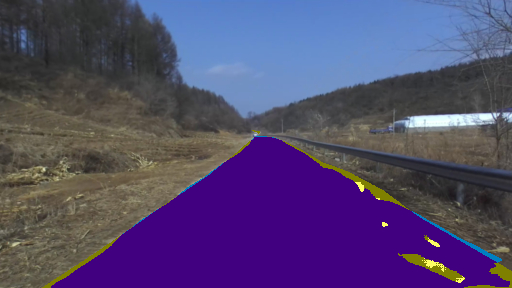}};
 \end{tikzpicture}
 \label{f048} 
 }

\vspace{0.02cm}

\subfloat{ 
 \centering 
 \begin{tikzpicture}
 \node[anchor=south west,inner sep=0] (image) at (0,0) {\includegraphics[width=0.122\textwidth]{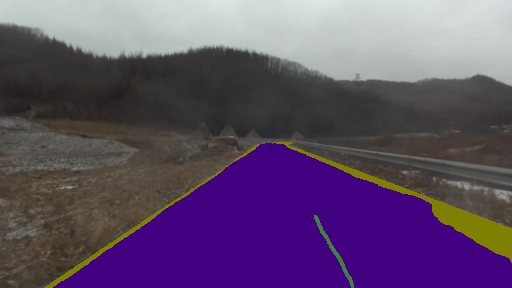}};
 \end{tikzpicture}
 \label{f051} 
 } \hspace{-0.33cm}
\subfloat{  
 \centering 
 \begin{tikzpicture}
 \node[anchor=south west,inner sep=0] (image) at (0,0) {\includegraphics[width=0.122\textwidth]{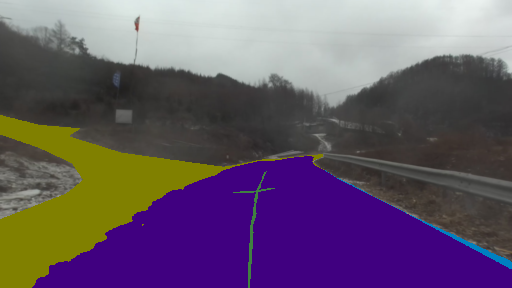}};
 \end{tikzpicture}
 \label{f052} 
 } \hspace{-0.33cm}
\subfloat{  
 \centering 
 \begin{tikzpicture}
 \node[anchor=south west,inner sep=0] (image) at (0,0) {\includegraphics[width=0.122\textwidth]{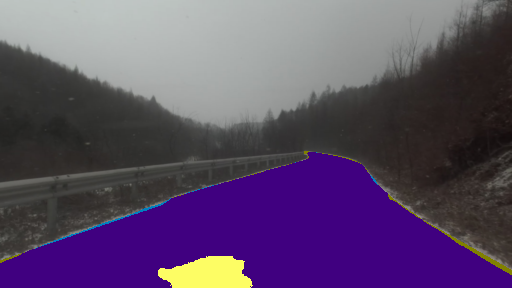}};
 \end{tikzpicture}
 \label{f053} 
 } \hspace{-0.33cm}
\subfloat{  
 \centering 
 \begin{tikzpicture}
 \node[anchor=south west,inner sep=0] (image) at (0,0) {\includegraphics[width=0.122\textwidth]{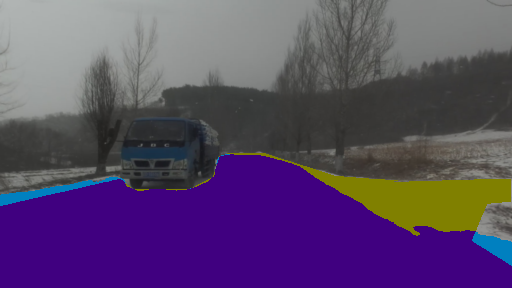}};
 \end{tikzpicture}
 \label{f054} 
 } \hspace{-0.33cm}
\subfloat{  
 \centering 
 \begin{tikzpicture}
 \node[anchor=south west,inner sep=0] (image) at (0,0) {\includegraphics[width=0.122\textwidth]{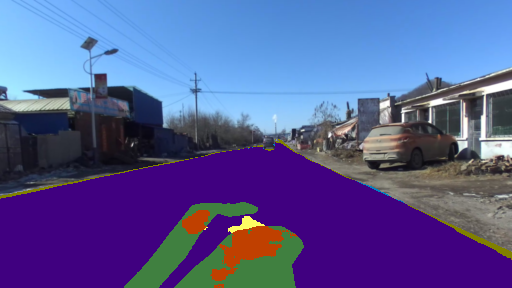}};
 \end{tikzpicture}
 \label{f055} 
 } \hspace{-0.33cm}
\subfloat{  
 \centering 
 \begin{tikzpicture}
 \node[anchor=south west,inner sep=0] (image) at (0,0) {\includegraphics[width=0.122\textwidth]{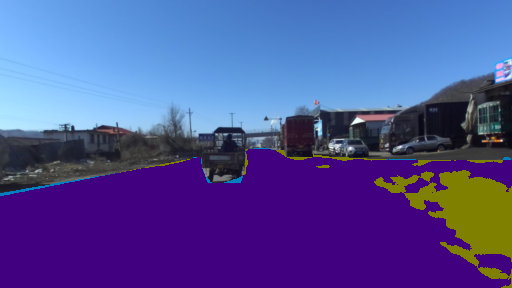}};
 \end{tikzpicture}
 \label{f056} 
 } \hspace{-0.33cm}
\subfloat{  
 \centering 
 \begin{tikzpicture}
 \node[anchor=south west,inner sep=0] (image) at (0,0) {\includegraphics[width=0.122\textwidth]{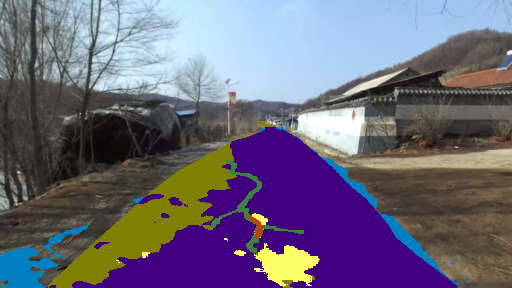}};
 \end{tikzpicture}
 \label{f057} 
 } \hspace{-0.33cm}
\subfloat{  
 \centering 
 \begin{tikzpicture}
 \node[anchor=south west,inner sep=0] (image) at (0,0) {\includegraphics[width=0.122\textwidth]{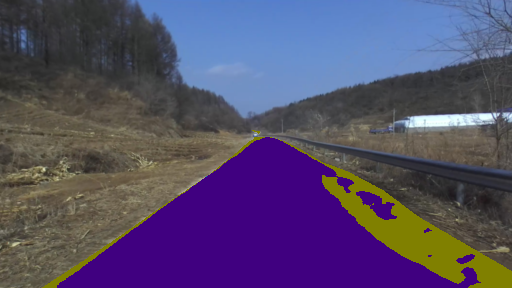}};
 \end{tikzpicture}
 \label{f058} 
 }

\vspace{0.02cm}

\subfloat{ 
 \centering 
 \begin{tikzpicture}
 \node[anchor=south west,inner sep=0] (image) at (0,0) {\includegraphics[width=0.122\textwidth]{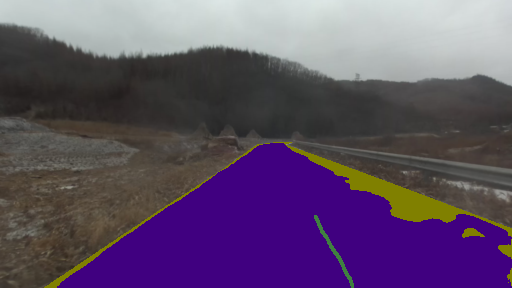}};
 \end{tikzpicture}
 \label{f061} 
 } \hspace{-0.33cm}
\subfloat{  
 \centering 
 \begin{tikzpicture}
 \node[anchor=south west,inner sep=0] (image) at (0,0) {\includegraphics[width=0.122\textwidth]{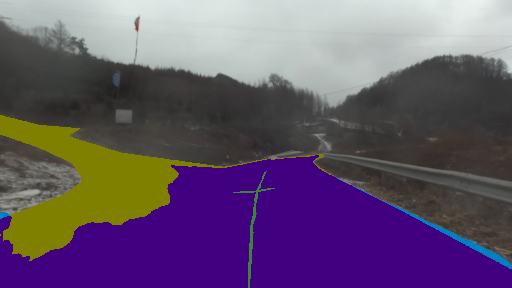}};
 \end{tikzpicture}
 \label{f062} 
 } \hspace{-0.33cm}
\subfloat{  
 \centering 
 \begin{tikzpicture}
 \node[anchor=south west,inner sep=0] (image) at (0,0) {\includegraphics[width=0.122\textwidth]{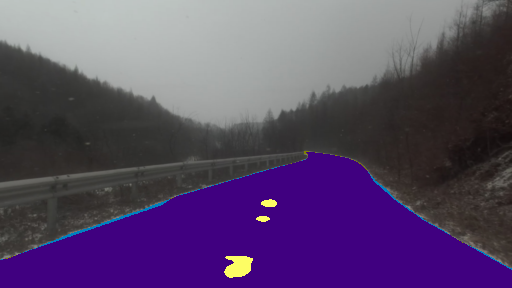}};
 \end{tikzpicture}
 \label{f063} 
 } \hspace{-0.33cm}
\subfloat{  
 \centering 
 \begin{tikzpicture}
 \node[anchor=south west,inner sep=0] (image) at (0,0) {\includegraphics[width=0.122\textwidth]{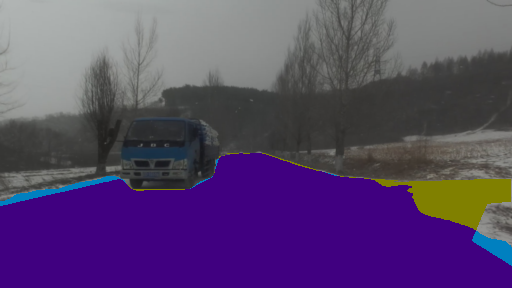}};
 \end{tikzpicture}
 \label{f064} 
 } \hspace{-0.33cm}
\subfloat{  
 \centering 
 \begin{tikzpicture}
 \node[anchor=south west,inner sep=0] (image) at (0,0) {\includegraphics[width=0.122\textwidth]{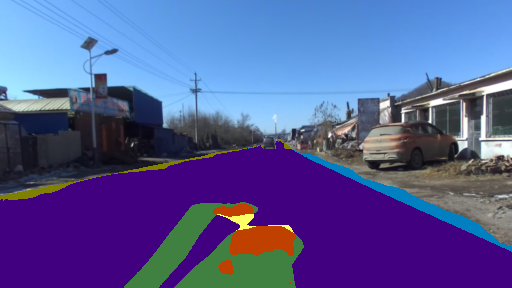}};
 \end{tikzpicture}
 \label{f065} 
 } \hspace{-0.33cm}
\subfloat{  
 \centering 
 \begin{tikzpicture}
 \node[anchor=south west,inner sep=0] (image) at (0,0) {\includegraphics[width=0.122\textwidth]{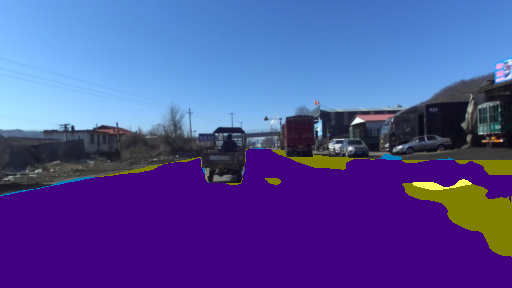}};
 \end{tikzpicture}
 \label{f066} 
 } \hspace{-0.33cm}
\subfloat{  
 \centering 
 \begin{tikzpicture}
 \node[anchor=south west,inner sep=0] (image) at (0,0) {\includegraphics[width=0.122\textwidth]{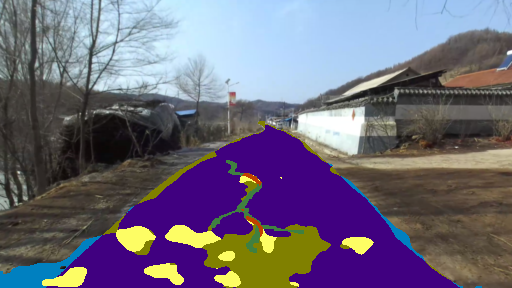}};
 \end{tikzpicture}
 \label{f067} 
 } \hspace{-0.33cm}
\subfloat{  
 \centering 
 \begin{tikzpicture}
 \node[anchor=south west,inner sep=0] (image) at (0,0) {\includegraphics[width=0.122\textwidth]{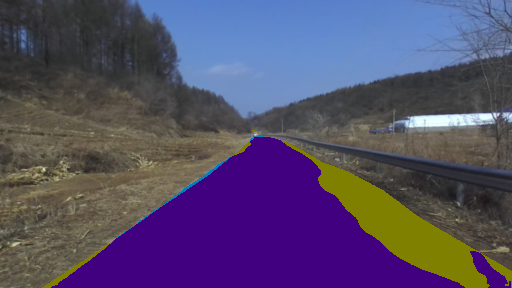}};
 \end{tikzpicture}
 \label{f068} 
 }
\caption{Sample qualitative demonstrations of the top-$3$ multi-modal fusion networks with the best mIoU metric. The $4$-th row to the $6$-th row are respectively demonstrations of our AMFNet, MAFNet \cite{feng2022mafnet}, and RoadSeg \cite{fan2020roadseg}. \textcolor{pothole}{$\mdblksquare$}, \textcolor{potholeFP}{$\mdblksquare$}, and \textcolor{potholeFN}{$\mdblksquare$} represent negative obstacles, the false positive of negative obstacles, and the false negative of negative obstacles. \textcolor{road}{$\mdblksquare$}, \textcolor{roadFP}{$\mdblksquare$}, and \textcolor{roadFN}{$\mdblksquare$} represent drivable roads, the false positive of drivable roads, and the false negative of drivable roads, respectively. The figure is best viewed in color.}
\label{results} 
\end{figure*}

Some sample qualitative results of the top-$3$ multi-modal fusion networks (i.e., our AMFNet, MAFNet, and RoadSeg) with the best mIoU metric are shown in Fig.~\ref{results}. From the third column of the results, we can see that the snow cover on the road confuses the segmentation of negative obstacles. MAFNet and RoadSeg incorrectly segment negative obstacles due to the influence of the snow cover. However, our AMFNet correctly segments the drivable road. From the results in the seventh column, we can see that the water and shadows on the road seriously degrade the segmentation performance of the drivable road. Our AMFNet also achieves the best results among the three networks. From the results in the fifth column, we can see that our AMFNet correctly segments most areas of negative obstacles. The results illustrate the superiority of our AMFNet.

\section{Conclusions and Future Work}

We proposed here a novel network AMFNet with the AMF module for the segmentation of drivable roads and negative obstacles. Our proposed network addresses the degradation of fusion performance caused by untrustworthy features extracted from depth images. We generated masks from depth images as the input of AMF module. The AMF module generates two adaptive-weight masks for the RGB feature map and depth feature map to reduce the influence of untrustworthy features. In addition, we released a large-scale RGB-D dataset with pixel-level ground truth of drivable roads and negative obstacles for this task. The experimental results show that our proposed AMFNet achieves better performance than single-RGB modality in the presence of untrustworthy features during fusion. However, there are also some limitations in our AMFNet. For example, the segmentation accuracy of the class \textit{negative obstacles} is low. We would like to improve the weight of the loss of the \textit{negative obstacles} during training in the future.

\section*{Acknowledgment}
This work was supported in part by the National Natural Science Foundation of China under Grant 62003286, in part by Zhejiang Lab under Grant 2021NL0AB01, and in part by the HK PolyU Start-up Fund under Grant P0034801.

\bibliographystyle{IEEEtran}
\bibliography{mybibfile}

\end{document}